\documentclass{mva_style}
\usepackage{graphicx}
\graphicspath{ {Figures/} }
\usepackage{amsmath}
\usepackage{comment}
\usepackage{subcaption}
\usepackage{romannum}
\usepackage{float}
\usepackage{xcolor}

\finalcopy 

\begin{document}
\title{Detecting Humans in RGB-D Data with CNNs}

\author{
  Kaiyang Zhou \\
  University of Bristol \\
  {\tt https://kaiyangzhou.github.io} \\
  \and
  Adeline Paiement \\
  Swansea University \\
  {\tt A.T.M.Paiement@swansea.ac.uk}\\
  \and
  Majid Mirmehdi \\
  University of Bristol \\
  {\tt m.mirmehdi@bristol.ac.uk}\\
}

\maketitle

\section*{\centering Abstract}
\textit{
We address the problem of people detection in RGB-D data where we leverage depth information to develop a region-of-interest (ROI) selection method that provides proposals to two color and depth CNNs. To combine the detections produced by the two CNNs, we propose a novel fusion approach based on the characteristics of depth images. We also present a new depth-encoding scheme, which not only encodes depth images into three channels but also enhances the information for classification. We conduct experiments on a publicly available RGB-D people dataset and show that our approach outperforms the baseline models that only use RGB data.
}

\section{Introduction}
RGB-D images encapsulate richer information by providing depth along with color values. In RGB-D human detection, depth information is usually used to reduce the search space \cite{massimo2016humanrgbd}. For example, Jafari et al. \cite{jafari2014real} use depth pixels to extract the regions of interest (ROIs) by classifying each pixel into one of three categories: ground plane, ROIs (objects) and non-ROIs (buildings and walls). Zhang et al. \cite{zhang2013real} remove the ground plane and ceiling, then propose ROIs based on the density distribution along the depth dimension. In this paper, we also leverage depth information to remove the ground plane, but we further constrain ROIs by exploiting the characteristics of depth.

Convolutional neural networks (CNNs), which exhibit significant powers of discrimination in the color image domain \cite{krizhevsky2012imagenet}, have been successfully applied to people detection. Angelova et al. \cite{angelova2015real} propose a cascade framework consisting of several CNNs for pedestrian detection where proposals are obtained by a dense sliding window. However, each CNN in each cascade stage is applied repeatedly to each proposal, without sharing computations on convolutions. We also use CNNs for proposal classification, but convolutions are performed only once on the entire image, which is achieved by the ROI-pooling proposed by Girshick \cite{girshick2015fast}. ROI-pooling extracts a fixed-size feature vector for each ROI window in the last set of convolutional feature maps and forwards these feature vectors to the fully connected layers, which actually require fixed-size input.

CNNs have also been applied in RGB-D images for human detection, such as \cite{martinson2016real,mees16iros}. In \cite{martinson2016real} however, depth is not included in the classification. Mees et al. \cite{mees16iros} develop a mixture of CNNs to extract features independently from different modalities including color, depth and motion, and use a gating network to fuse them for further classification. However, they use the traditional sliding-window approach to generate proposals, which ignores the potential of depth for ROI selection. In this paper, we apply two CNNs to learning features from color and depth images respectively. They perform human detection in ROIs of each modality separately, followed by a fusion of these results to obtain a more reliable detection overall.

Our contributions in this paper are: (a) we develop a fast ROI selection method based on depth to reduce the search space, (b) we propose a CNN-based RGB-D human detector where we design a novel way to fuse detections from RGB and depth images, (c) we propose a fast depth-encoding method, which can produce three-channel depth images that are close to color images in terms of saliency of information. This allows a more effective deployment of depth information by CNNs through the transfer of pre-learnt color features.

\section{Proposed Approach} \label{sec:approach}

\begin{figure}[t]
\centering
\includegraphics[width=75mm]{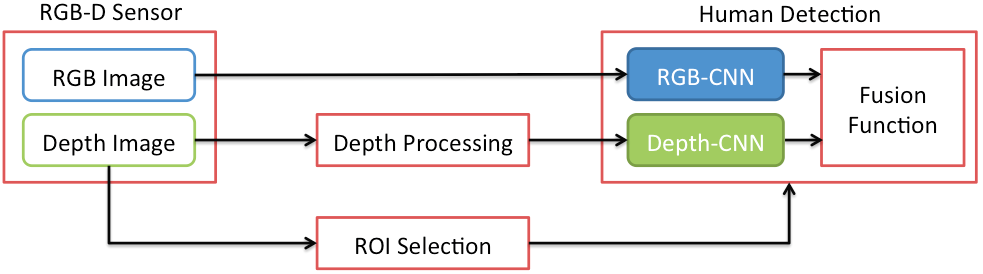}
\caption{Overview of our human detection system.}
\label{fig:detectpipeline}
\end{figure}

Fig. \ref{fig:detectpipeline} shows a compact overview of our human detection system. The {ROI Selection} module exploits the depth information to reduce the search space and generates a set of candidate proposals. The {Depth Processing} module fills holes in the depth image using a mean-filter, and then it encodes the filled image into three channels. The two networks, RGB-CNN and Depth-CNN, in the {Human Detection} module, process the RGB and encoded depth images respectively, producing probability scores for the candidate proposals. For each proposal, two probability scores produced by the two networks are combined by a fusion function, leading to a stronger probability that indicates whether the proposal contains a person (upper body) or not. In the end, proposals with high probabilities are kept and then passed through non-maximum suppression, resulting in the final list of non-overlapping windows.

\subsection{Depth Processing} \label{sec:depthprocess}
A good depth-encoding method needs to satisfy three criteria: (i) it should preserve as much information contained in the original depth as possible, such as shape, (ii) it should be computationally cheap, and (iii) it should produce an image with characteristics that matches that of a color image, notably in terms of range and contrast. Based on these, we evaluate three existing methods, namely {\itshape depth-gray} (DG) and {\itshape color-depth} (CD) from \cite{eitel2015multimodal} and {\itshape contrast-enhanced depth-gray} (CE) from \cite{crabbe2015skeleton}, and propose a new scheme in this work. In particular, DG normalizes depth pixels to have values ranging from 0 to 255 and then replicates them to three channels. CD maps each normalized pixel to three channels via a reversed jet colormap\footnote{https://uk.mathworks.com/help/matlab/ref/colormap.html} where the resulting color values are ranged from red (near) through green to blue (far). CE is similar to DG, except that it performs histogram equalization after the normalization and before the replication into the three channels. We further propose a new depth-encoding scheme called {\itshape contrast-enhanced color-depth} ({CECD}), which performs histogram equalization before mapping pixels to the reversed jet colormap. These four schemes are compared in Section \ref{sec:experiment}.

\subsection{ROI Selection} \label{sec:ourROIselection}
The ROI Selection process is composed of three stages, outlined below,  which work together to produce candidate proposals.
\footnote{The code for the ROI selection module is available on https://github.com/KaiyangZhou/ROI-Selection}

\begin{figure}[h]
\centering
\begin{center}
  \begin{subfigure}[b]{0.37\linewidth}
  \includegraphics[height=2.5cm]{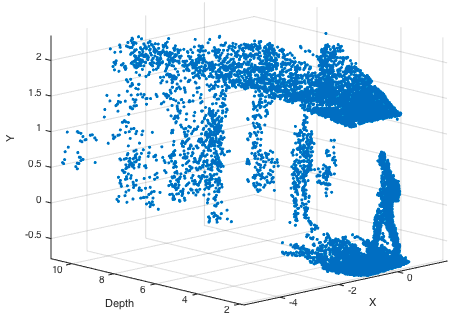}
  \caption{} \label{fig:3dworld}
  \end{subfigure}
  \hfill
  \begin{subfigure}[b]{0.37\linewidth}
  \includegraphics[height=2.5cm]{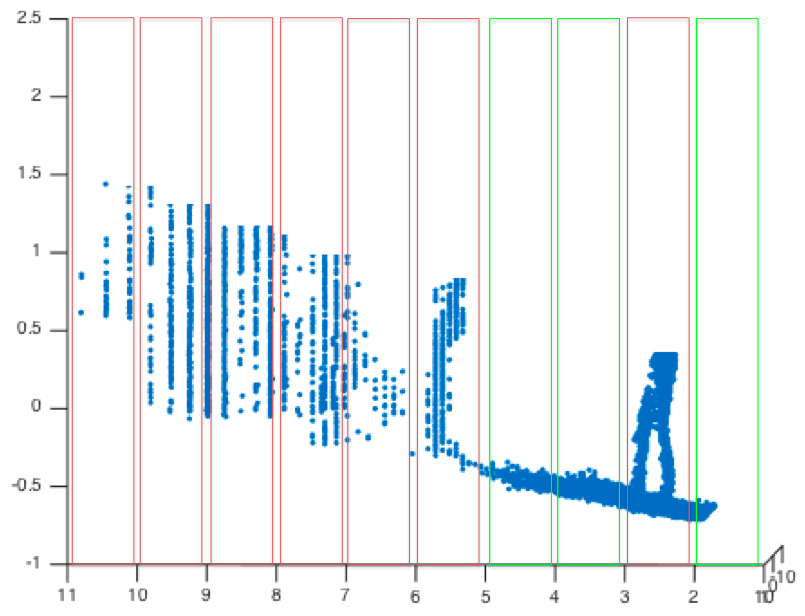}
  \caption{} \label{fig:sideview}
  \end{subfigure}
  \hfill
  \begin{subfigure}[b]{0.22\linewidth}
  \includegraphics[height=2.5cm]{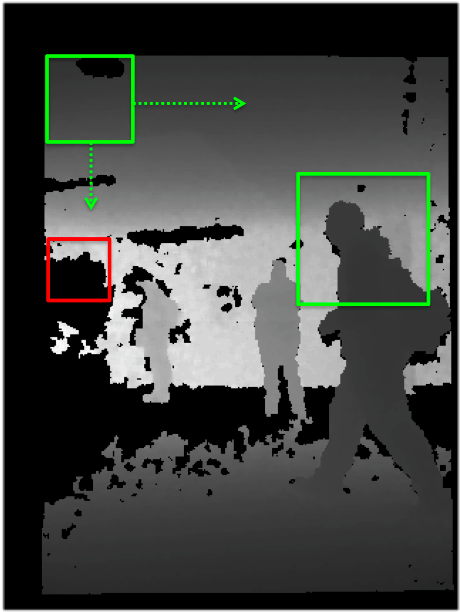}
  \caption{} \label{fig:proposalfiltering}
  \end{subfigure}
\end{center}
\caption{ROI selection - see text for explanation.}
\end{figure}

\textbf{Ground Plane Detection (GPD) -} We project depth pixels to the global 3D world with known depth camera intrinsic parameters (see Fig.~\ref{fig:3dworld}). To determine the ground plane accurately in the presence of numerous outliers, we sample the 3D world into a grid of $10\times10$ cells (see Fig.~\ref{fig:sideview}). Since the ground is usually located in the lower part of a scene, we only sample points from the lower half of the image. For each bin, we compute standard deviation of the points in vertical columns spanning this area (VSTD). The bins with VSTD values larger than an empirical threshold are removed and the rest of the points are fed to a RANSAC-based plane fitting algorithm \cite{yang2010plane}. This is based on the observation that bins with points largely spread in the vertical direction will contain fewer ground pixels and thus should be filtered out. In the end, pixels close to the plane correspond to non-ROIs and the rest are selected for further ROI selection. Jafari et al. also use an outlier-reduction method before detecting the ground plane in \cite{jafari2014real}. However, they remove bins with a high density of points, which can lead to the removal of bins that may contain pixels that belong to the ground plane at locations close to the camera (in this case, the points are quite compact).

\textbf{Scale-Informed ROI Search (SIS) -} We slide a window across the remaining pixels. The window width for each pixel is dynamically determined by the depth value which has the benefit of {avoiding a time-consuming multi-scale search}. In particular, the window width $\lambda$ at a specific pixel is determined by $\lambda = \frac{fW}{Z}$, where $f$ is the focal length of the depth camera, $W$ is the rough width of a normal human (we used 0.6m), and $Z$ is the depth value (in meters). We only detect the human upper body, so each window is a square bounding box.

\textbf{Candidate Proposals Filtering (CPF) -} To further reduce the number of proposals, we discard those that mainly contain invalid pixels (no depth value) arising from objects with poor reflective properties. To efficiently compute the number of valid pixels in a bounding box, we employ the idea of integral images \cite{viola2001rapid}. A proposal is only selected if its portion of valid pixels is larger than a threshold e.g. one-third. This process can be executed in parallel with the SIS stage to keep the computation fast. Fig.~\ref{fig:proposalfiltering} shows an example of valid (green) and invalid (red) proposals.

\subsection{Human Detection with CNNs} \label{sec:peopledetect}
\begin{figure}[t]
\centering
\includegraphics[width=\linewidth]{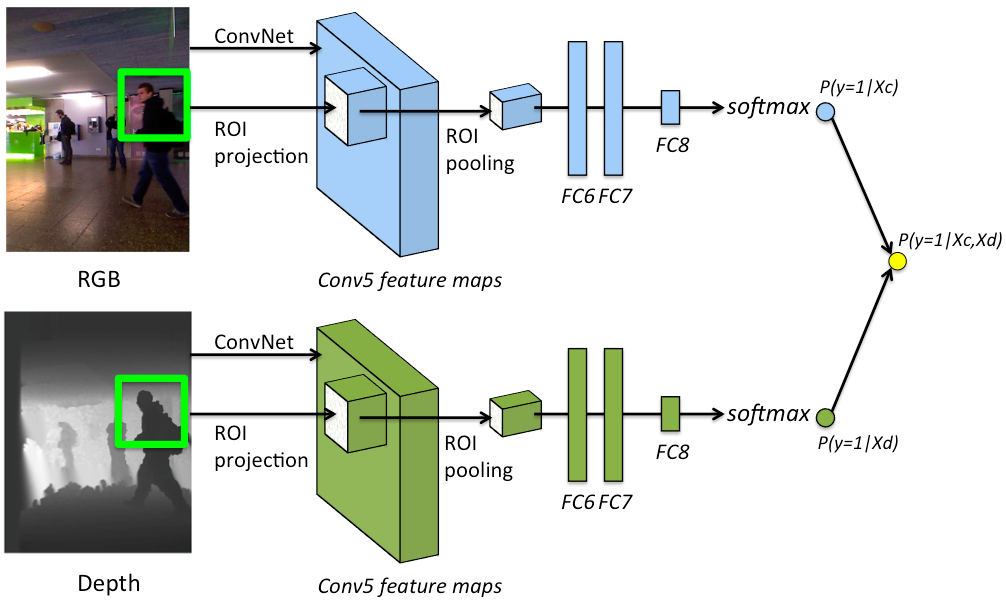}
\caption{RGB-D human detection CNN architectures. The blue and green streams process RGB and depth images respectively. Each network takes as input a full-size image and the same set of ROIs. The detection probabilities from each network are fused to infer a stronger probability.}
\label{fig:peopledetector}
\end{figure}

The architecture of our system is shown in Fig.~\ref{fig:peopledetector}. The blue stream network (RGB-CNN) processes RGB images while the green stream network (Depth-CNN) processes depth images. They are identical in architecture but do not share parameters. We use  CaffeNet  \cite{jia2014caffe} but change the 1000-way classification layer with a 2-way classification layer to suit our purposes. CaffeNet is essentially a variant of AlexNet \cite{krizhevsky2012imagenet} where the order of pooling and normalization layers is switched. We replace the max-pooling layer after the $5^{th}$ convolutional layer with the ROI-pooling layer \cite{girshick2015fast}. Further, we follow \cite{girshick2015fast} to restructure the FC6 and FC7 layers via truncated SVD, resulting in size-reduced weight parameters which helps increase the processing speed. We apply multi-scale detection at test time \cite{girshick2015fast}.

{\bf Fusion of RGB and Depth Detections - }
The two networks work independently in each domain to score each proposal. For each proposal, the RGB-CNN produces a probability $P(y=1|X_c)$ and the Depth-CNN produces a probability $P(y=1|X_d)$, where $y=1$ means a window contains a person (upper body), and $X_c$ and $X_d$ represent the proposal region in RGB and depth images respectively. The two probabilities are fused to deduce the final probability $P(y=1|X_c,X_d)$, which is a stronger evidence for classifying the proposal. We compute the fusion probability as
\begin{equation} \label{eq:fusergbd}
\scriptsize
P(y=1|X_c,X_d) \propto exp((1-\omega) L(y=1|X_c) + \omega L(y=1|X_d)) ,
\end{equation}
where $L(\cdot | \cdot)$ is the log likelihood and $\omega$ is an adaptive weight dependent on the proposal depth, defined as
\begin{equation} \label{eq:weightfunction}
\scriptsize
\omega = \begin{cases}
1 & d <= 1\\
-\frac{1}{5}(d-1)+1 & 1 < d < 6\\
0 & d >= 6\\
\end{cases}  ~,
\end{equation}
where $d$ is the depth value in meters and its threshold values in (\ref{eq:weightfunction}) are set empirically. In Eq.~(\ref{eq:fusergbd}),  if the proposal is close to the camera, we give more weight to the depth information, while if the proposal is far away from the camera, we rely more on RGB information. This is grounded in  the fact that the performance of most depth sensors downgrades as the distance increases. It is also easy to observe in depth images that people closer to the camera have more clear shapes than people who are far away (see e.g. Fig.~\ref{fig:proposalfiltering}).

\section{Experiments} \label{sec:experiment}
\textbf{Dataset -}
For training data, we collect positive training data provided by SPHERE \cite{tao2015comparative} and negative training data from NUYD2 dataset \cite{silberman2012indoor}. This results in 3271 positive images and 7574 negative images. We choose the {\it RGBD people dataset} \cite{spinello2011people} to evaluate our approach. This dataset contains three videos each including 1000+ RGB-D images. We found that some annotations were missing in this subset, thus we manually added any missing annotations.\footnote{The completed annotation set is available on https://github.com/KaiyangZhou/new-annotations-rgbdpeople}

\textbf{Network Training -}
We use {\it Caffe} \cite{jia2014caffe} to train and test our networks with the SGD algorithm (learning rate 0.001, momentum 0.9 and weight decay 0.0005). All the networks are initialized with ImageNet weights, except for the new 2-way classification layer which is initialized with a Gaussian distribution. Fine-tuning is applied to the RGB-CNN (only for fc layers) for 3 epochs where the learning rate is lowered by $1/10$ after 2 epochs. The Depth-CNN is fine-tuned with different settings. For each depth-encoding scheme (see Section \ref{sec:depthprocess}), the one producing the best result when combining with the RGB-CNN is selected for later comparison.\footnote{The best fine-tuning for the four Depth-CNNs are:  (1) DG-CNN:  fc678, epoch=1; (2) CE-CNN: conv345fc678, epoch=3; (3) CD-CNN: conv345fc678, epoch=3; (4) CECD-CNN: fc678, epoch=1.}

\textbf{Evaluation Methodology -}
For evaluation, we determine average-precision against average-recall. We adopt the `no-reward-no-penalty' rule \cite{spinello2011people} with positive detections at 50\% overlap and penalize repeated detections on the same person. We use three models as the baseline: RCNN \cite{girshick2014rich} + our ROI method, our RGB-CNN + our ROI method, and our RGB-CNN + SelectiveSearch \cite{uijlings2013selective}. We disable the bounding box regression in RCNN as we found that our proposal method is less sensitive to localization error. We tailor SelectiveSearch particularly for upper body detection by setting the height of each proposal to its width, keeping only the square upper part of elongated proposals, and discarding proposals with width $<50$ pixels. The first two baselines let us assess the advantage of using depth information for the detection stage, as well as our proposed color and depth fusion method. The third baseline allows the evaluation of our ROI method comparing to SelectiveSearch used in \cite{girshick2014rich}.

\begin{figure}[t]
\centering
\includegraphics[width=\linewidth]{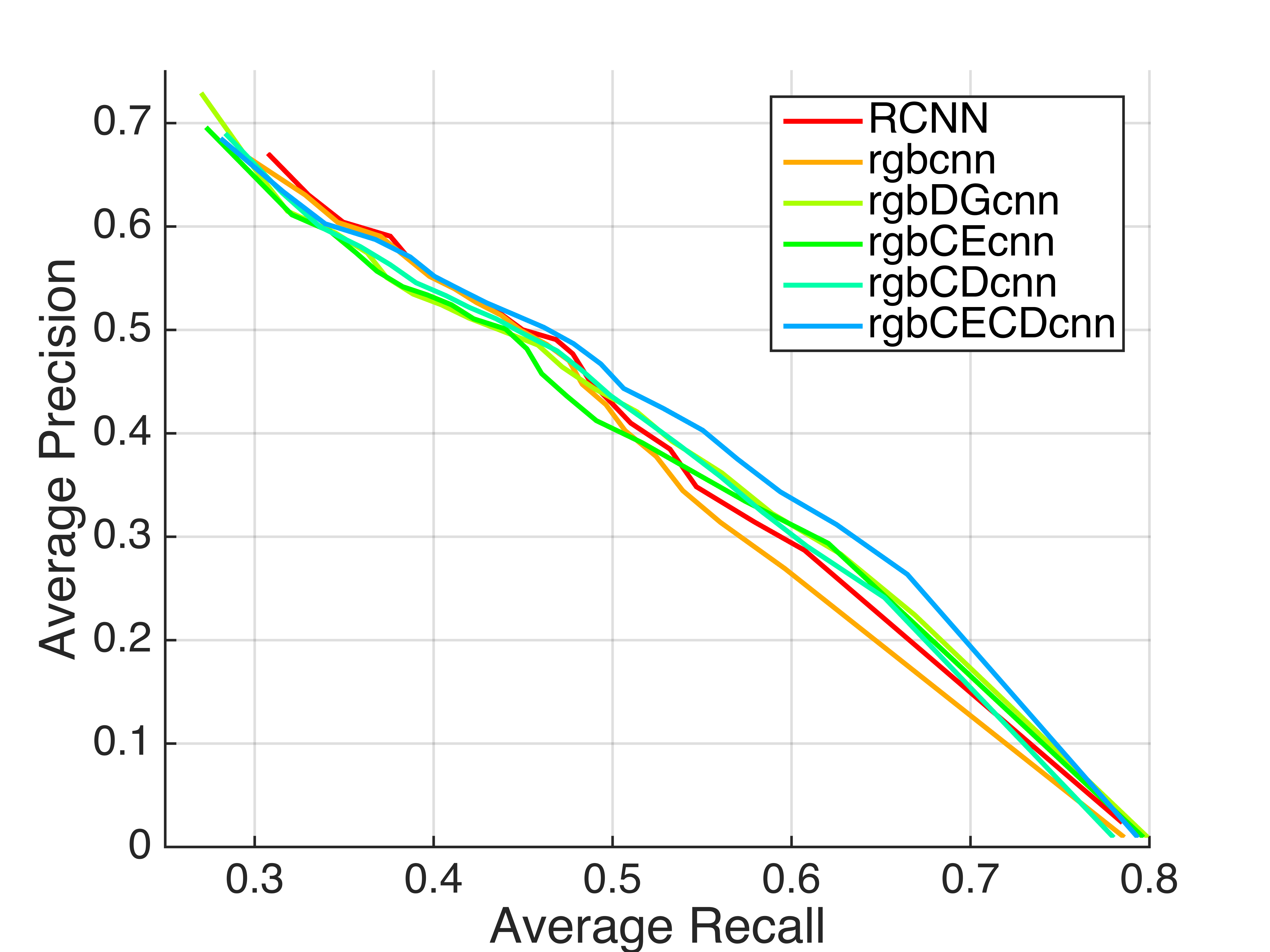}
\caption{Experimental results of different models and using different depth encodings. The model with the CECD encoding (i.e. rgbCECDcnn) performs the best.}
\label{fig:new_exp}
\end{figure}

\textbf{Results and Discussion -}
The experimental results are shown in Fig.~\ref{fig:new_exp} where rgbcnn represents our RGB-CNN and rgb$*$cnn with $* \in \{ \text{DG, CE, CD, CECD} \}$ represents combined RGB-CNN and Depth-CNN with different depth encodings.

{\it Depth encoding -} Overall, the performances of the depth encoding schemes, used with the same pre-trained CNN, can be sorted in the decreasing order as: CECD $>$ CD $>$ DG $>$ CE for a targeted average recall of 0.5. These results show that the CECD-encoding is better at producing new depth images close in characteristics to color images. In other words, by using the CECD-encoding, the fine-tuning of the Depth-CNN requires less effort in adjusting the pre-trained weights.

\begin{figure*}[t]
\centering
  \begin{subfigure}[t]{0.155\textwidth}
  \includegraphics[width=\textwidth]{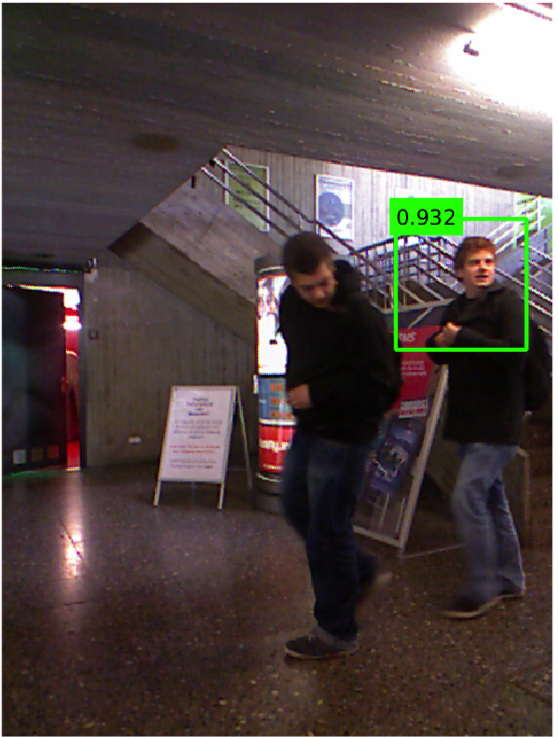}
  \caption{} \label{fig:rgbonly1}
  \end{subfigure}
  \begin{subfigure}[t]{0.155\textwidth}
  \includegraphics[width=\textwidth]{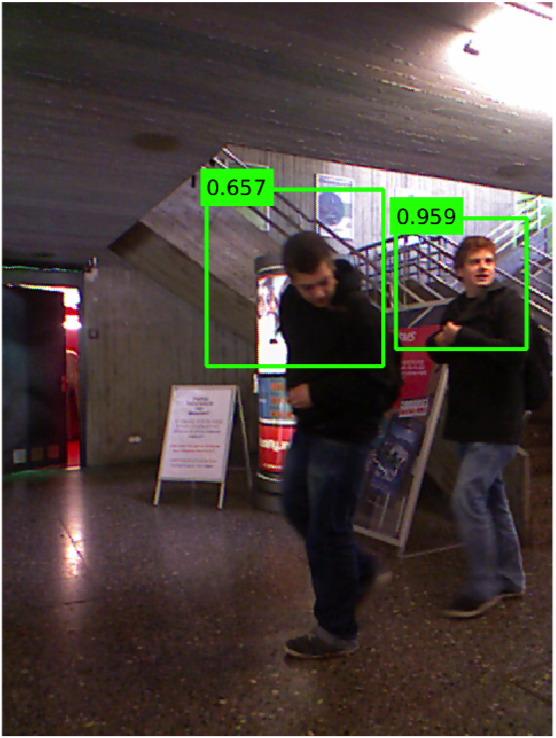}
  \caption{} \label{fig:rgbdepth1}
  \end{subfigure}
  \begin{subfigure}[t]{0.155\textwidth}
  \includegraphics[width=\textwidth]{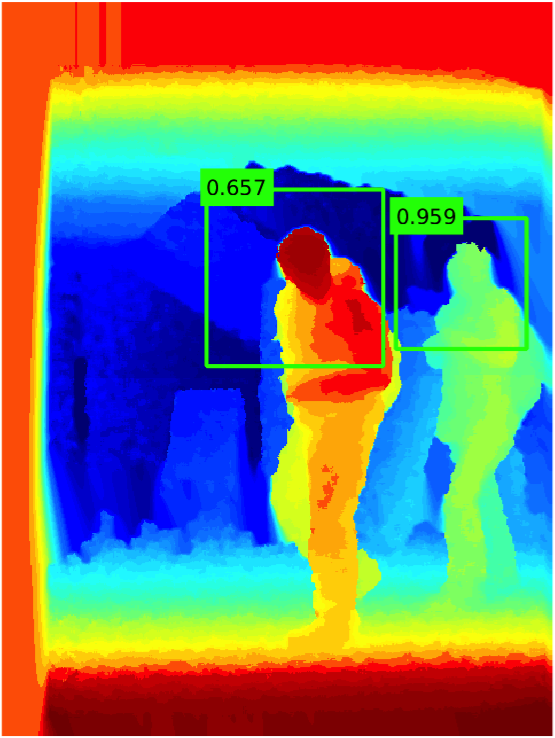}
  \caption{} \label{fig:depth1}
  \end{subfigure}
  \begin{subfigure}[t]{0.155\textwidth}
  \includegraphics[width=\textwidth]{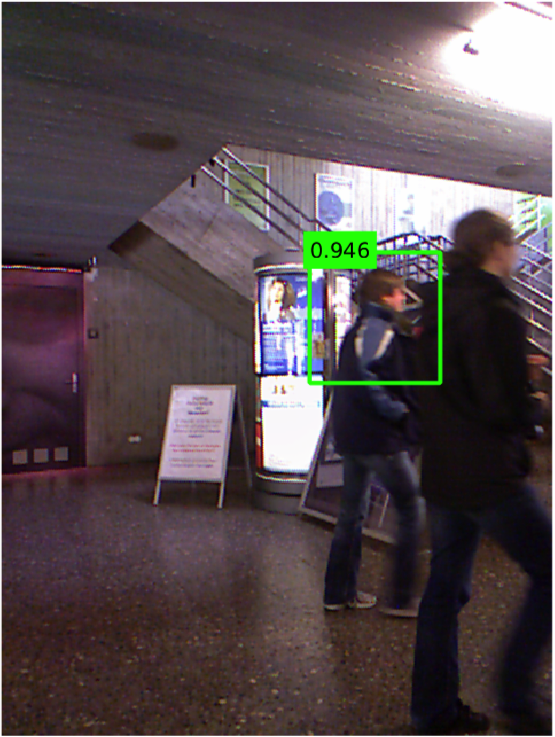}
  \caption{} \label{fig:rgbonly2}
  \end{subfigure}
  \begin{subfigure}[t]{0.155\textwidth}
  \includegraphics[width=\textwidth]{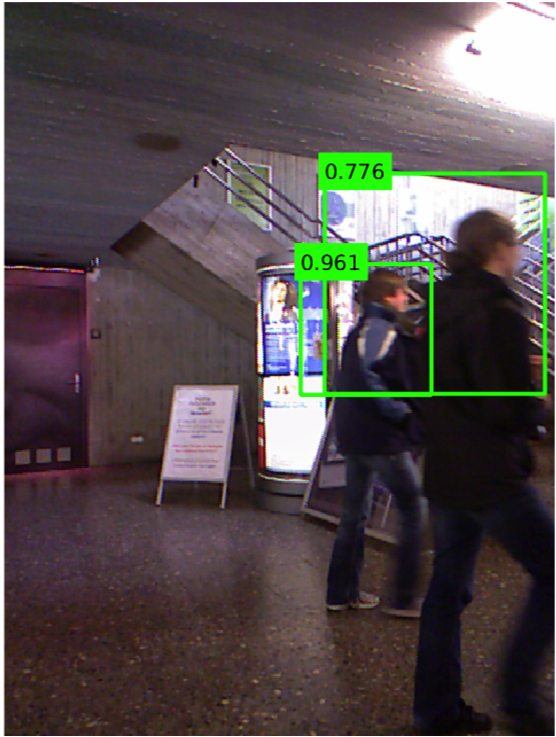}
  \caption{} \label{fig:rgbdepth2}
  \end{subfigure}
  \begin{subfigure}[t]{0.155\textwidth}
  \includegraphics[width=\textwidth]{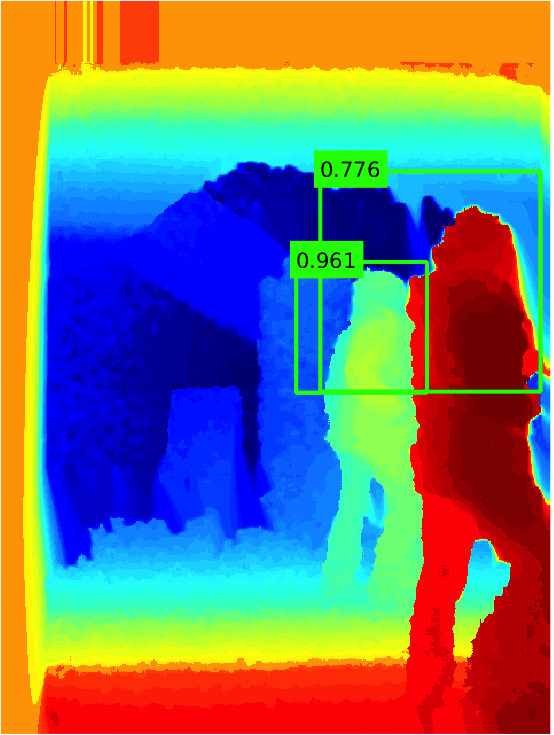}
  \caption{} \label{fig:depth2}
  \end{subfigure}
  \caption{Detection results obtained by rgbcnn (a,d) and rgbCECDcnn (b,c,e,f). It can be seen that the shape/contrast information is well preserved in the CECD depth images (c) and (f).}
  \label{fig:detectresults}
\end{figure*}

{\it Combined RGB and depth detection -} 
The rgbCECDcnn outperforms the two baselines, showing that by adding depth the detection accuracy can be improved. Two observations may explain this result: (i) \textit{Depth-CNN is more robust to pose deformations.} The front person in Fig.~\ref{fig:rgbonly1} is a missing detection when using only RGB-CNN. This might be caused by the slightly unusual body pose. By combining with Depth-CNN, this person can be detected (Fig.~\ref{fig:rgbdepth1}), as the shape of the person is still a discriminating feature in the encoded depth image (Fig.~\ref{fig:depth1}). (ii) \textit{Depth detections can compensate the absence of RGB detections when color information is not discriminative enough.} When using only RGB-CNN, the front person in Fig.~\ref{fig:rgbonly2} is missed. This is probably because the appearance of the person is somewhat blurred and the skin information is insufficient. Fortunately, in the (encoded) depth domain (Fig.~\ref{fig:depth2}), the strong shape and contrast information can compensate these deficiencies. 

\begin{figure}[h]
\centering
\includegraphics[width=\linewidth]{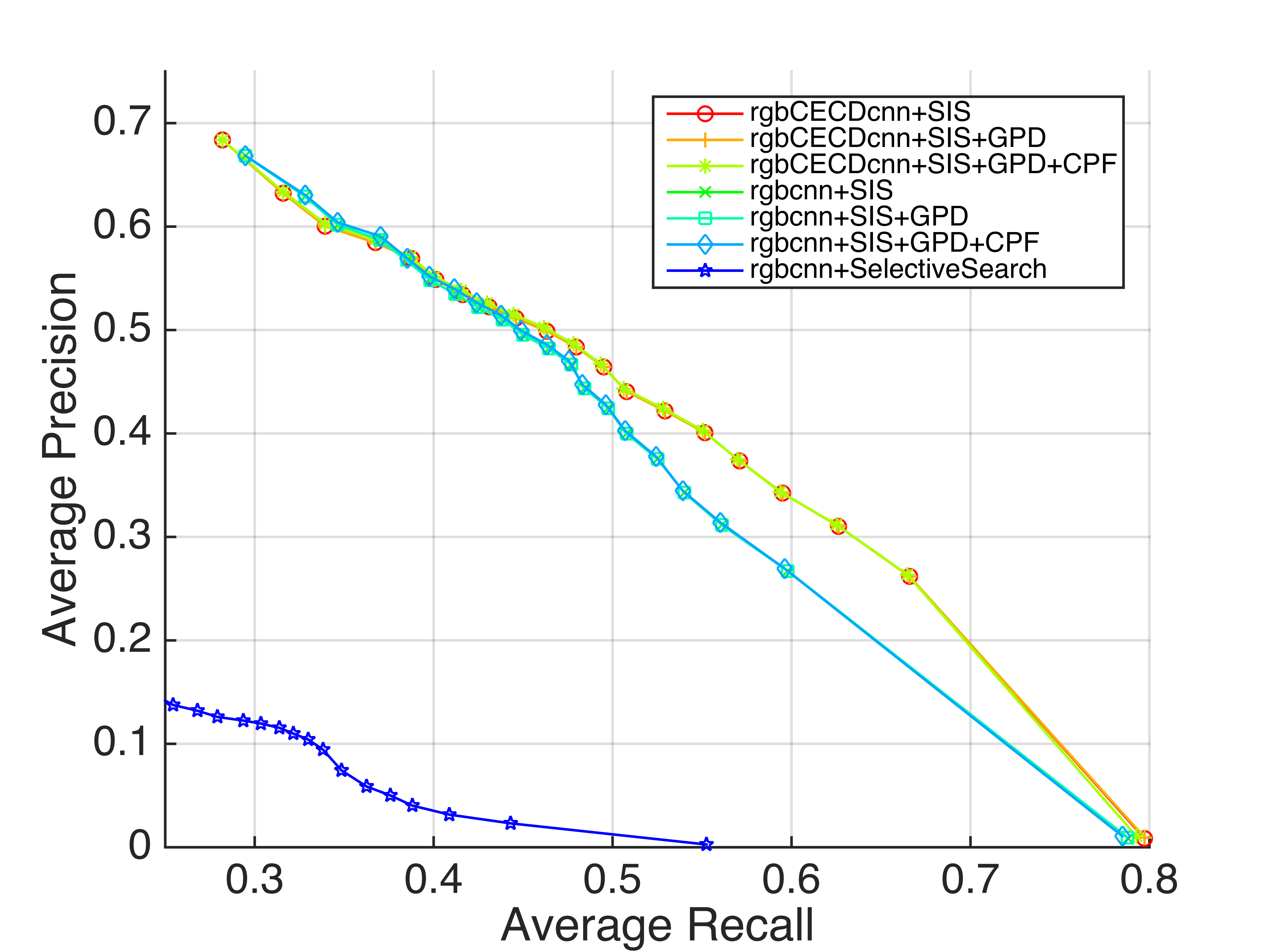}
\caption{Evaluation of the ROI selection method.}
\label{fig:ROIperform}
\end{figure}

{\it Depth based ROI selection -}
Fig.~\ref{fig:ROIperform} shows that for each CNN model, the results of using SIS, SIS+GPD and SIS+GPD+CPF nearly overlap, which strongly suggests that the speed improvement gained by our ROI method does not significantly affect our detection accuracy. The selective search result exposes a severe low-precision problem, with many false positives or repeated detections. Our ROI method did not suffer from these problems, partly because SIS prevents producing large windows on far backgrounds and small windows on close foregrounds. As for speed, although our ROI method produces more proposals (5000 vs. 1500 with tiny proposals being pruned as mentioned), the time spent is two orders of magnitude shorter than that of selective search (14.5ms vs. 2700ms).

\section{Conclusion and Future Work} \label{sec:conclude}
With the infiltration of RGB-D cameras on the one hand and the potential for fast and powerful CNN-based solutions on the other in the vision community, solutions to the problem of people detection in RGB-D data with CNNs are in demand. In this paper, we proposed (1) an effective ROI selection method based purely on depth, (2) a depth-encoding method and (3) a two-stream CNNs framework for people detection, including a novel color-depth fusion approach. We demonstrated that by combining color and depth detections our models outperformed the RGB baselines. In the future, it would be interesting to learn the fusion function parameters from data, which will allow us to perform end-to-end training on the two networks.

\bibliographystyle{ieeetr}
\bibliography{Bib}

\end{document}